\documentclass[letterpaper, 10 pt, conference]{ieeeconf}  

\IEEEoverridecommandlockouts                              

\usepackage{color}
\newcommand{\ie}{i.e.\ }

\newcommand{\Reffig}[1]{Figure~\ref{#1}}
\newcommand{\Refsec}[1]{Section~\ref{#1}}

\newcommand{\Refeq}[1]{Equation~\ref{#1}}
\newcommand{\Reftab}[1]{Table~\ref{#1}}

\usepackage{graphics} 
\usepackage{epsfig} 

\usepackage{algorithm}
\usepackage{algpseudocode}
\usepackage{amsmath}
\usepackage{amssymb}
\usepackage{subfigure}
\usepackage{graphicx}
\usepackage{subfigure}
\usepackage{booktabs}
\usepackage{threeparttable}
\usepackage{multirow}
\usepackage{multicol}
\usepackage{makecell}
\usepackage{bm}
\usepackage{makecell}

\title{\LARGE \bf
Scale Estimation with Dual Quadrics for Monocular Object SLAM
}
\author{Shuangfu Song$^{2}$, Junqiao Zhao$^{*, 1}$, Tiantian Feng$^{2}$, Chen Ye$^{1}$, Lu Xiong$^{3}$
\thanks{*This work is supported by the National Key Research and Development Program of China (No. 2021YFB2501104), the National Natural Science Foundation of China (No. 41871370)}
\thanks{$^{2}$Shuangfu Song and Tiantian Feng are with the School of Surveying and Geo-Informatics, Tongji University, Shanghai 200092, China
        {\tt\footnotesize (e-mail: 1911204@tongji.edu.cn; fengtiantian@tongji.edu.cn).}}
\thanks{$^{1}$Junqiao Zhao and Chen Ye are with the Department of Computer Science and Technology, School of Electronics and Information Engineering, Tongji University, Shanghai 201804, China
        {\tt\footnotesize (e-mail: zhaojunqiao@tongji.edu.cn; yechen@tongji.edu.cn).}}
\thanks{$^{3}$Lu Xiong is with the Institute of Intelligent Vehicle, Tongji University, Shanghai, 201804 China
        {\tt\footnotesize(e-mail: xiong\_lu@tongji.edu.cn).}}%
}

\begin{document}

\maketitle
\thispagestyle{empty}
\pagestyle{empty}

\begin{abstract}
The scale ambiguity problem is inherently unsolvable to monocular SLAM without the metric baseline between moving cameras. 
In this paper, we present a novel scale estimation approach based on an object-level SLAM system. 
To obtain the absolute scale of the reconstructed map, we formulate an optimization problem to make the scaled dimensions of objects conform to the distribution of their sizes in the physical world, without relying on any prior information about gravity direction.
The dual quadric is adopted to represent objects for its ability to describe objects compactly and accurately, thus providing reliable dimensions for scale estimation.
In the proposed monocular object-level SLAM system, semantic objects are initialized first from fitted 3-D oriented bounding boxes and then further optimized under constraints of 2-D detections and 3-D map points.
Experiments on indoor and outdoor public datasets show that our approach outperforms existing methods in terms of accuracy and robustness.
\end{abstract}

\section{INTRODUCTION}

Recently, object-level SLAM has achieved significant progress thanks to breakthroughs in convolutional neural networks (CNNs).
A series of monocular object SLAM systems \cite{nicholson2019QuadricSLAM,yang2019CubeSLAM,wu2020EAOSLAM} are proposed.
These semantically enriched SLAM algorithms can improve robot intelligence for scene understanding and human-robot interaction.
However, due to the limitation of monocular sensors, the absolute scale between the reconstructed map and the physical world is unknown, which is known as the scale ambiguity problem.
This largely hinders the application of these algorithms in actual scenarios.
Hence, it is crucial and necessary to estimate the absolute scale in monocular SLAM systems.


For monocular cameras, the absolute scale is unobservable but can be inferred from prior knowledge such as the sizes of objects in the scene \cite{konkle2012RealWorld}.
Based on this insight, some previous studies \cite{sucar2017Probabilistic,sucar2018Bayesian} attempted to use the Bayesian framework to infer the absolute scale by associating estimated heights of observed objects with their priors.
However, there are some limitations:
First, when measuring the height of an object, it has to resort to the direction of gravity as the reference.
Objects that are not upright are treated as anomalous.
Second, only specific objects with prominent height can be used in the algorithm, other low objects such as ``keyboard'', ``book'', etc., are discarded.
As a result, the applicability and accuracy of these methods are limited.

In this paper, we propose a novel absolute scale estimation approach by exploring all reliable dimensions of general objects without resorting to the prior gravity direction.
Therefore, more constraints from general objects can reduce the uncertainty of scale estimation and make the scaled dimensions of objects closer to those of the physical world.

To obtain accurate dimension estimation of objects for scale inferencing, we develop a monocular object SLAM system using the dual quadric as the three-dimensional (3-D) object representation.
A dual quadric can flexibly fit various 3-D shapes tightly with compact parameters, thus the size of the object can be well measured.
Furthermore, quadrics in dual vector space are well defined in projective geometry \cite{Hartley2003} and can be robustly initialized \cite{chen2021Robust}.
We first initialize the objects with a two-stage strategy based on an improved version of our previous work \cite{chen2021Robust}.
Then accurate dimensions can be extracted from dual quadrics reconstructed during semantic mapping. 
The Metric-tree \cite{zhang2020Scaleaware}, a repository of sizes of more than 900 object categories, is used to provide object size priors. 
We further introduce uncertainties to Metric-tree priors to ensure that the result will not be skewed by the prior sizes with large deviation.
At last, an optimization problem is established to estimate the absolute scale factor.

In summary, our contributions are as follows:
\begin{itemize}
    \item[1] An accurate and robust scale estimation approach exploiting dual quadrics and object size priors.
    \item[2] A monocular object SLAM system with a semantic mapping module, which can build object-oriented maps accurately and timely.
\end{itemize}
To the best of our knowledge, this is the first work that takes advantage of dual quadrics to estimate the scale factor within a monocular object SLAM system.

\section{RELATED WORK}

\subsection{Scale Estimation}
To overcome the monocular scale ambiguity problem, an intuitive way is to leverage extra sensors, such as inertial measurement units (IMUs) \cite{nutzi2011Fusion}, LiDAR \cite{zhang2018Scale}.
Without augmented sensors, at least one absolute reference needs to be integrated.
Reference information used in existing approaches can be divided into three categories: camera setup parameters, learned models, and prior information of the semantic object.

On ground-based vehicle platforms, cameras are usually mounted at a specific height above the ground plane.
\cite{song2014Robust, zhou2016Reliable,wang2018Monocular} proposed to estimate unscaled camera height by extracting the ground plane parameters derived from map points.
Then, with the known camera height relative to the ground plane, the absolute scale can be obtained.
These approaches are accurate and effective, but cannot cover more general scenarios, for example, where only unanchored hand-hold cameras are available.

In \cite{yin2017Scale}, a monocular SLAM system was proposed to recover the scale by incorporating a depth prediction module based on deep convolutional neural fields. 
A similar idea is proposed in DVSO \cite{yang2018Deep}, in which deep depth predictions are used as virtual stereo measurements.
The difference is that \cite{yang2018Deep} trained the network with sparse depth reconstructions from Stereo DSO \cite{wang2017Stereo}, instead of using depth measurements captured by a LiDAR.
Another learning-based method \cite{greene2020Metrically} trained a space-coarse, but depth-accurate CNN for depth prediction, and directly estimated the scale factor for every frame. 
These approaches have achieved impressive performance on KITTI datasets.
However, it is still a challenge for them to generalize to different environments.

To exploit the semantic object priors, methods in \cite{sucar2017Probabilistic} and \cite{sucar2018Bayesian} adopt the Bayesian framework incorporating the height priors of objects to infer the global scale.
Recent work \cite{zhang2020Scaleaware} improved this approach by introducing not only heights but all available dimensions of objects into the scale estimation.
However, these approaches all require the gravity direction known and objects to be placed upright to measure the size of objects. 

\subsection{Object SLAM}
Classic visual SLAM systems have been well explored in representing geometric primitives like points, lines, and planes.
SLAM++ \cite{salas2013SLAM} is considered to be the first object-level SLAM system that uses RGB-D detection and represents objects by matching prior models.
Similarly, \cite{galvez2016Real} leveraged a large object database and uses bags of words to identify objects.
Later, more general object representations were adopted to break through the restriction of the pre-prepared object database.
\cite{frost2018Recovering} represented objects as spheres and incorporated them into bundle adjustment as extended points.
QuadricSLAM \cite{nicholson2019QuadricSLAM} was proposed to represent objects flexibly using dual quadrics.
CubeSLAM \cite{yang2019CubeSLAM} described objects using cuboids.
Inspired by them, EAO-SLAM \cite{wu2020EAOSLAM} adopted both quadrics and cuboids to represent objects based on their different prior shapes.
However, cuboids require additional constraints, such as uprightness, for the reconstruction, as its projection is not well defined in the projective geometry. 

\section{ABSOLUTE SCALE ESTIMATION} \label{sec_scale}

\subsection{Problem Formulation}
We use the ellipsoid, a closed surface form of the dual quadric, to describe a general 3-D object.
It has nine parameters: 6 DoF pose and three dimensions \ie{length, width and height}.
We denote $d$ as a unscaled dimension of reconstructed objects, $\tilde{d}$ as its real length.
The absolute scale factor $s$ can be expressed as
\begin{equation}
    s = \frac{\tilde{d}}{d}.
\end{equation}
Then the reconstructed map can be corrected by this scale.
However, each dimension in each object can calculate a different local scale factor through its prior length, which is globally inconsistent;
This requires us to use all dimensions of all objects as conditions to find a global optimal scale.

We assume that each dimension of general objects is conforming to a Gaussian distribution $N(\mu,\sigma)$, and dimensions are independent of each other.
With the set of all dimensions $D = \{d_i\}$ of objects, the conditional probability of scale $s$ is presented as
\begin{equation}
    P(s|D) \propto \prod_i {P(d_i|s)}.
\end{equation}
Our goal is to find the scale which maximizes the probability according to the distribution of dimensions in the physical world.
The likelihood can then be presented as 
\begin{equation}
    P(d_i|s) = N(\mu_i,\sigma_i)
    \label{eq_prior_scale}
\end{equation}
where $\mu_i$ and $\sigma_i$ are prior parameters to describe the distribution of $d_i$ in the physical world.
The error between scaled $d_i$ and its expection is defined as 
\begin{equation}
    e_i = \mu_i - sd_i.
\end{equation}
The maximum likelihood estimation (MLE) can be formulated as a least-squares optimization problem
\begin{equation}
    s^* = \mathop{\arg\min}\limits_{s} \sum_i \left\lVert \frac {e_i} {\sigma_i} \right\rVert ^2.
    \label{opti scale}
\end{equation}

\subsection{Object Uncertainty Model}

The uncertainty of a dimension, as shown in \Refeq{eq_prior_scale}, can be modeled by size statistics from the Metric-tree \cite{zhang2020Scaleaware}.
However, the accuracy of the scale estimate is also affected by the quality of the object reconstruction.
We define a confidence $c$ value to evaluate the reliability of an object based on observations used for dual quadrics construction:
\begin{equation}
    c = \frac{w_1c_{det} + w_2c_{pt} + w_3c_{vis}}{w_1 + w_2 + w_3}
\end{equation}
where $w_1,w_2,w_3$ are weight parameters.

$c_{det}$ is the confidence of 2-D detection and is derived by:
$$
  c_{det}=\frac{1}{n}\sum_k p_k
$$
where $p_k$ is the probability of detection obtained from the 2-D object detector in view $k$.

$c_{pt}$ is the confidence derived from $N_p$, the total number of map points associated with the object: 
$$
c_{pt} = \min(1,\max(0, \log_{a} N_p)).
$$

$c_{vis}$ is the confidence of visibility and is derived by: 
$$
c_{vis} = \min(1,\max(0, \log_{b} N_o))
$$
where $N_o$ is the total number of 2D detections.
$a,b$ are super parameters set to be 10 and 15 respectively in our implementation.

Lastly, \Refeq{opti scale} can be reformulated by integrating the confidence weight as:
\begin{equation}
    s^* = \mathop{\arg\min}\limits_{s} \sum_i \left\lVert \frac {c_i e_i} {\sigma_i} \right\rVert ^2.
    \label{opti scale2}
\end{equation}

\subsection{Object Dimensions Selection}
Besides the uncertainty, the reliability of the dimensions of an object should also be taken into account.
For example, dimensions like the height of ``bottle", length of ``spoon" are more stable than the thickness of ``book” and ``keyboard”.
This is because small dimensions are difficult to estimate accurately during the mapping process.
Therefore, objects are classified into three categories according to their dimensions.
We denote three dimensions of a object as $d_1, d_2, d_3$ following $d_1 \geq d_2 \geq d_3 > 0$.
Then the shape feature of an object can be defined as: 
\begin{equation}
    L_d = \frac{d_1-d_2}{d_1} \quad 
    P_d = \frac{d_2-d_3}{d_1} \quad 
    S_d = \frac{d_3}{d_1}
\end{equation}
where $L_d$, $P_d$, $S_d$ are linearity, planarity, and scattering, respectively \cite{weinmann2014Semantic}.
For an object with $S_d<0.3$, it belongs to the pole-like when $L_d>0.5$; it belongs to the disk-like object when $P_d>0.5$.
For a pole-like object, only the longest dimension of it is stable enough to be used.
For a disk-like object, the shortest dimension of it should be discarded.
For others, all dimensions can provide useful hints for scale inferencing. 

\subsection{Outliers Elimination}
False 2-D object detections lead to the erroneous association between 3D objects and their corresponding size priors, therefore should be eliminated. 

We found that most local scale factors estimated are consistent with each other.
Therefore, outliers caused by false detection can be detected and eliminated by statistical methods such as the boxplot.
First, all local scales are sorted in ascending order.
Then the interquartile range (IQR) is defined as $\mathrm{IQR} = Q_3 - Q_1$.
where $Q_1$ is the first quartile and $Q_3$ is the third quartile.
A dimension will be discarded if its local scale factor is less than $Q_1-1.5\mathrm{IQR}$ or greater than $Q_3+1.5\mathrm{IQR}$.

\subsection{Implementation}
\begin{figure}
    \centering
    \includegraphics[width=\linewidth]{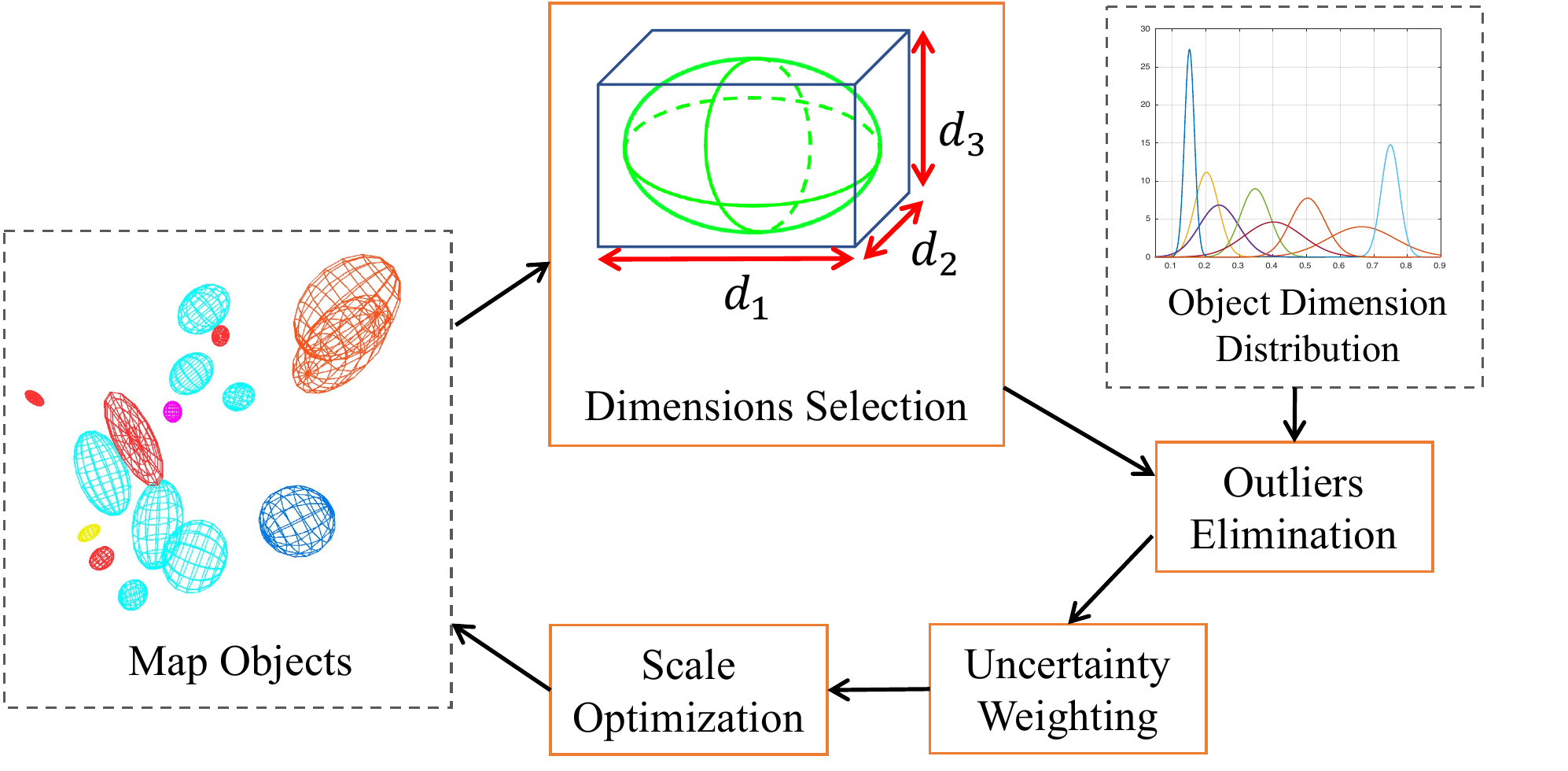}
    \caption{The pipeline of scale estimation.}
    \label{fig_scale}
\end{figure}

Our scale estimation pipeline is shown in \Reffig{fig_scale}.
We first select all stable dimensions from map objects and calculate their local scales with prior dimension distribution for outliers elimination.
Then we combine the prior variance of the dimension and the confidence of the reconstructed object to weight each error term of the scale optimization process.
Scale estimation is embedded in the back-end optimization process of our SLAM system as described in \Refsec{sec_mapping}.
Every time the object map is updated, a new scale factor will be calculated automatically by g2o \cite{kummerle2011G2o}, and then be used to scale the whole map.

\section{SEMANTIC MAPPING} \label{sec_mapping}
\begin{figure}
    \centering
    \includegraphics[width=0.8\linewidth]{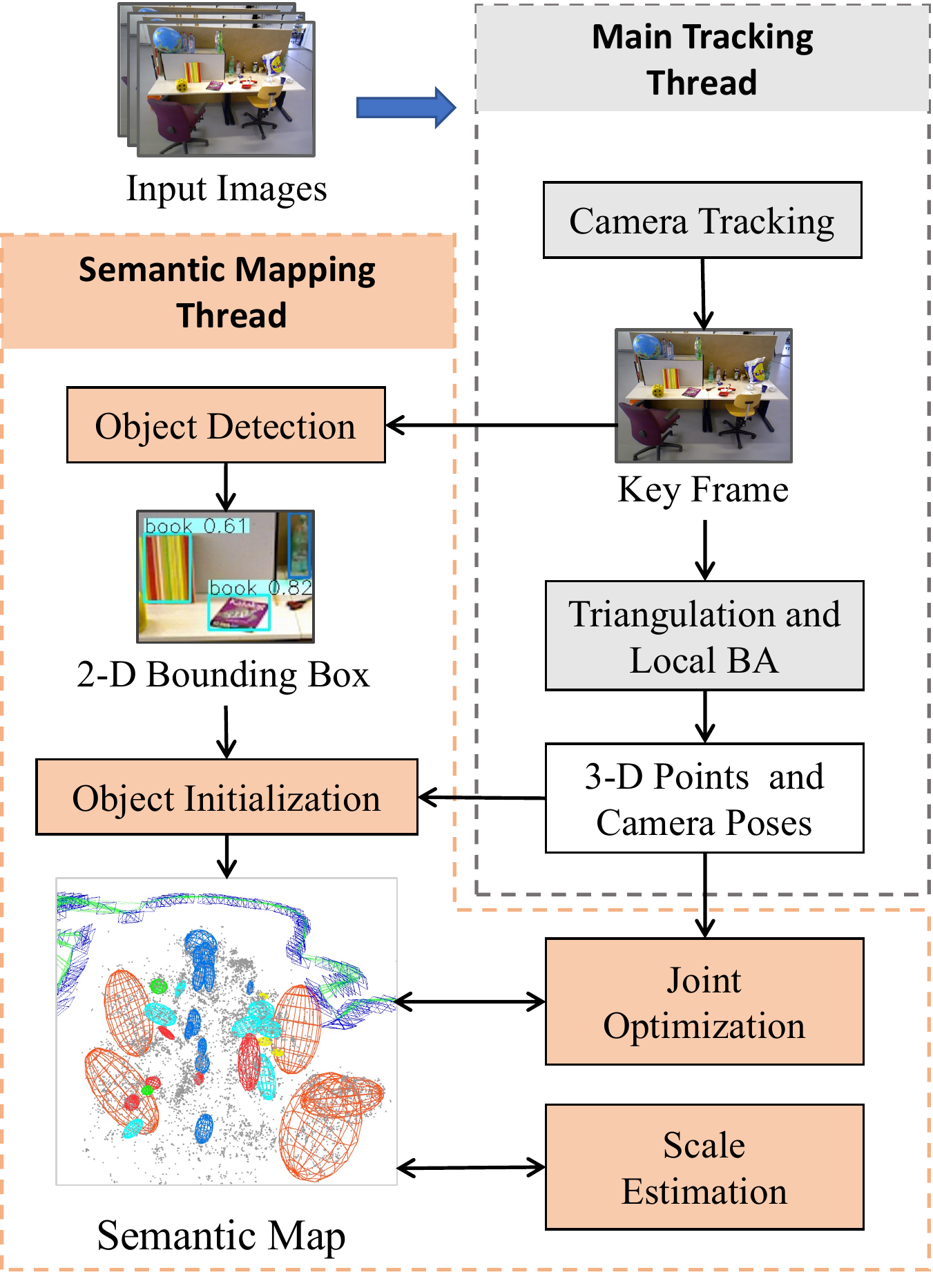}
    \caption{The proposed object SLAM system framework.}
    \label{fig_slam}
    \vspace{-3pt}
\end{figure}

\subsection{Overview}
To acquire accurate object size for scale estimation, we build an object SLAM system based on ORB-SLAM2 \cite{mur-artal2017ORBSLAM2}.
Only keyframes created by camera tracking are used for object detection.
Then dual quadrics are initialized from 2-D bounding boxes and 3-D points in the semantic mapping thread.
At last, all objects, points and camera poses are jointly optimized to update the semantic map.
The framework of the system is shown in \Reffig{fig_slam}.


\subsection{Dual Quadric Initialization} \label{sec_qua_ini}
A 3-D dual quadric can be initialized from a set of planar constraints constructed by multi-view detections for this object \cite{nicholson2019QuadricSLAM}.
However, this method only works when enough detections with diverse viewpoints are accumulated.
Our previous work \cite{chen2021Robust} constructed a 3-D convex hull from map points to provide valid planar constraints for dual quadric initialization.
In this paper, we propose an efficient 3-D oriented bounding box (OBB)-based approach for dual quadric initialization.

We ﬁrst associate 3-D map points to their corresponding object if points are projected into the 2-D bounding box of this object.
These map points are then clustered to filter out background points similar to \cite{chen2021Robust}.
Afterwards, OBB is extracted from associated map points to directly provide initial parameters (orientation, position, size) for dual quadrics.
Here, the adopted covariance-based OBB fitting algorithm is proposed in \cite{gottschalk2000collision}.
With the accumulation of measurements, the dual quadric will be re-initialized from 2-D detections as done in \cite{chen2021Robust}.

\subsection{Joint Optimization}
After initialization, objects are further optimized jointly with other map components.
Denote the set of camera poses, dual quadric object and points as $ X =\{x_i \in SE(3)\}$, $Q=\{q_j \in \mathbb{R}^{4 \times 4}\}$ and $P=\{p_k \in \mathbb{R}^3\}$, respectively.
Three types of measurement errors are introduced into the bundle adjustment.
\subsubsection{Camera-Object Measurement Error}
A dual quadric, represented by a $4\times4$ symmetric matrix $Q^*$, can be projected onto an image plane to obtain a dual conic represented by $3\times3$ symmetric matrix $C^*$, following this rule:
\begin{equation}
    C^* = HQ^*H^{\mathrm{T}}
\end{equation}
where $H = K[R|\bm{t}]$ is the camera projection matrix composed of camera intrinsic and extrinsic parameters.
As done in \cite{ok2019Robust}, we can obtain the predicted 2-D bounding box $\bm{\hat{b}} = [\hat{u}_{max}, \hat{v}_{max}, \hat{u}_{min}, \hat{v}_{min}]$ of the dual quadric by 
\begin{align}
    \hat{u}_{max}, \hat{u}_{min} &= \frac{1}{C^*_{3,3}} \left( C^*_{1,3} \pm \sqrt{ {C^*_{1,3}}^2 - C^*_{1,1} C^*_{3,3}} \right)\\
    \hat{v}_{max}, \hat{v}_{min} &= \frac{1}{C^*_{3,3}} \left( C^*_{2,3} \pm \sqrt{ {C^*_{1,3}}^2 - C^*_{2,2} C^*_{3,3}} \right)
\end{align}
where $[u_{max},v_{max}], [u_{min},v_{min}]$ represent the top left and bottom right corners of the 2-D box, respectively.
Denote the 2-D bounding box measurement observed by the object detector as $\bm{b}$.
Given a camera pose $x$, and a 3-D object $q$, the 4-D re-projected error can be defined as 
\begin{equation}
    \bm{e}(x,q) = \hat{\bm{b}} - \bm{b}.
\end{equation} 
\subsubsection{Camera-Point Measurement Error}
We use the standard 3-D point reprojection error, the same as in ORB-SLAM2 \cite{mur-artal2017ORBSLAM2}:
\begin{equation}
    \bm{e}(x,p) = \pi(T_c^{-1}p)-z
\end{equation}
where $\pi(.)$ is the camera projective function, $T_c$ is the camera pose, and $z$ is the pixel coordinate measurement.

\subsubsection{Object-Point Measurement Error}
\begin{figure}
    \centering
    \includegraphics[width=0.5\linewidth]{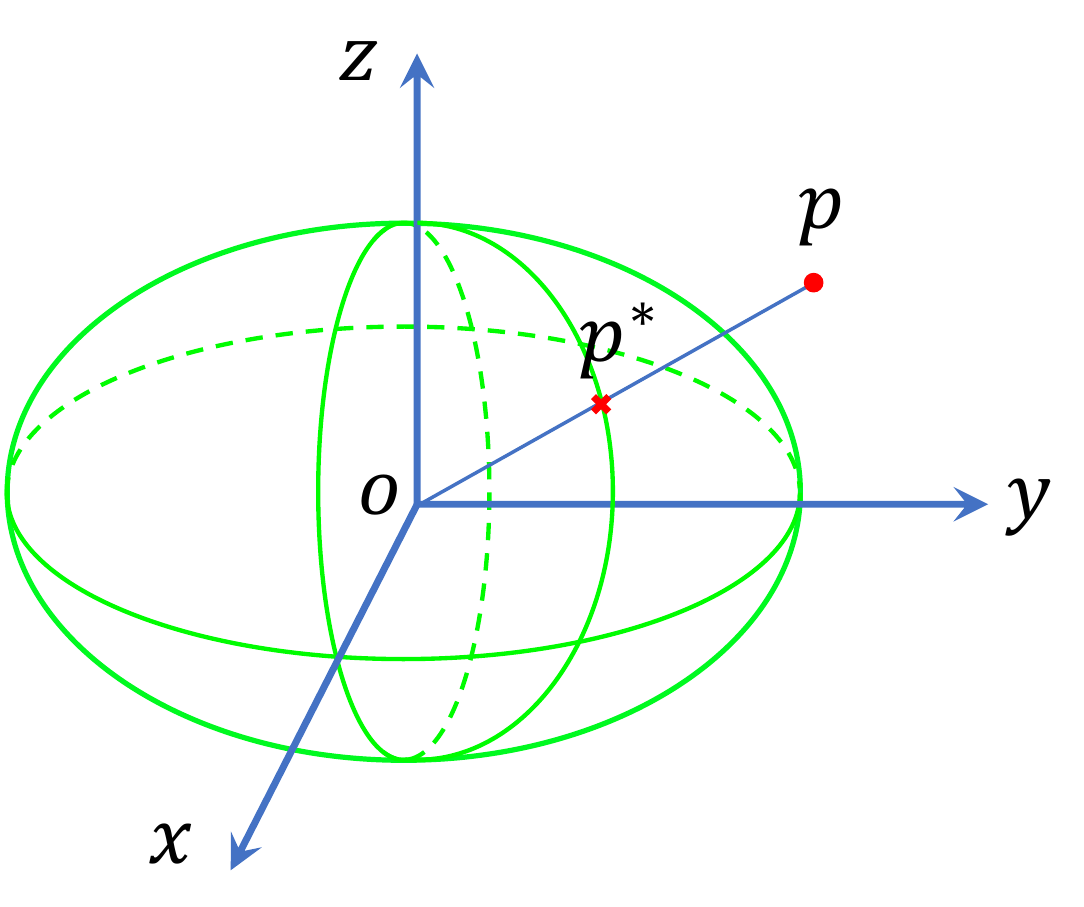}
    \caption{The measurement error between quadrics and points.}
    \label{fig_errors}
\end{figure}

To exploit constraints between map objects and map points, we introduce a novel measurement error:
\begin{equation}
    \bm{e}(q,p) = \max(0,\sqrt{p^{\mathrm{T}}Qp+1}-1)
\end{equation}
where $Q$ is the primal quadric matrix with $Q_{4,4} = -1$.
As shown in \Reffig{fig_errors}, when point $p$ is outside the ellipsoid, $p^*$ is the intersection point between the line segment $\overline{op}$ and the quadric surface.
the error represents the ratio $\overline{op^*}/\overline{p^*p}$.
We use the $\mathrm{max}$ operator to encourage the quadric to wrap the associated map points.
When a point lies inside the quadric, its measurement error will be zero.

Lastly, combining all measurement error items, the optimization problem can be formulated as
\begin{equation}
\begin{split}
    & X^*,Q^*,P^* =  \mathop{\arg\min}\limits_{\{X,Q,P\}} \big\{ \sum_{i,j} \left\lVert \bm{e}(x_i,q_j) \right\rVert_{\Omega_{ij}}^2 \\
    &  + \sum_{i,k}\left\lVert \bm{e}(x_i,p_k) \right\rVert_{\Omega_{ik}}^2 + \sum_{j,k}\left\lVert \bm{e}(q_j,p_k) \right\rVert_{\Omega_{jk}}^2 \big\}
\end{split}
\end{equation}
where $\Omega$ is the covariance matrix of different error measurements for Mahalanobis norm.
This nonlinear least-squares problem can be solved efﬁciently using Levenberg-Marquardt algorithm.
\section{EXPERIMENTS}
We evaluate the performance of our scale estimation approach on the TUM RGB-D data \cite{sturm2012Benchmark} as an indoor dataset and the KITTI tracking sequence \cite{geiger2012Are} as an outdoor dataset.
For monocular object SLAM, only RGB images are input and YOLOv3 \cite{redmon2018YOLOv3} are adopted for 2-D object detection.
We adopt the relative scale error (RSE) as the evaluation metric of scale estimation:
\begin{equation}
    \mathrm{RSE} = \frac{|\hat{s}-s| }{s}
\end{equation}
where $\hat{s}$ is our estimated global scale factor of the scene, and $s$ is the ground truth scale factor which can be obtained from the similarity transformation $S=\{s,R,\bm{t}\}$ between the trajectory estimated by our SLAM algorithm and the ground truth trajectory as following:
\begin{equation}
    S^* = \mathop{\arg\min}\limits_{\{s,R,\bm{t}\}} \sum_i \left\lVert p_i-s R \hat{p}_i-\bm{t} \right\rVert ^2
\end{equation}
where $\hat{p}_i$, $p_i$ represent the position of estimated trajectory, groundtruth trajectory respectively.

For the indoor dataset with common objects, we use the size information provided by the Metric-tree \cite{zhang2020Scaleaware}, which is collected from the internet, and construct a size distribution for each class of objects.
For the outdoor dataset, only the ``car'' object is used as reference, and its prior size distribution is based on the statistics of KITTI 3-D object annotations.
The remaining few classes of objects are either rarely seen in the scene or difficult to initialize successfully and are therefore discarded.

\subsection{Scale Estimation Results}

There are two works \cite{sucar2017Probabilistic,zhang2020Scaleaware} similar to ours.
The former first measures the heights of observed objects and then integrates the corresponding height priors into a Bayesian framework to infer the absolute scale.
The latter uses a structure-from-motion (SFM) system running offline to reconstruct 3-D scenes and a pixel-wise instance segmentation method to extract objects from the reconstructed point cloud, which is difficult to compare fairly with our approach.
Therefore, the former is chosen as our baseline.
Both methods use identical object size priors and a scale consistency-based outlier rejection algorithm. 
We selected three sequences from the TUM dataset and six sequences from the KITTI dataset for evaluation.
The selection criteria were two-fold: first, there should be representative semantic landmarks in the sequence; second, our SLAM system could run well on the sequence.
We ran each algorithm ten times on all sequences to calculate the average RSE and its standard deviation.

As shown in \Reftab{tb_rse_tum} and \Reftab{tb_rse_kitti}, our method achieves an average RSE of around 5\% and 10\% on TUM and KITTI sequences respectively, which is better than the baseline \cite{sucar2017Probabilistic}.
In the KITTI dataset, our method outperforms the baseline in most cases, except for the sequences 0000 and 0014, which are very short and have a limited number of objects available for scale estimation (from 2 to 4).

In the fr1\_desk sequence, the baseline method performed not well because there are few objects with stable height to use as reference, as shown in the far left of \Reffig{fig_qualitative_proj}.
Our method effectively utilizes the size information of ``book'' and ``keyboard'' in this scene to obtain more accurate and robust results.

In the other two sequences of TUM, our method performs better despite the fact that the baseline method has more object height cues. 
This is because the baseline method measures object height based on 2D detection results, which means it is susceptible to object occlusion.
Meanwhile, our method exploits multi-view observations of the object to reduce the uncertainty of dimension measurement, which improves the performance of scale estimation.

In addition, the scale estimation in the fr3-office sequence is slightly worse than in the other two scenes.
The main reason is that the dimension of some bottles in this scene exceeds the regular size in the Metric-tree \cite{zhang2020Scaleaware}.

In KITTI long sequences (except 0000 and 0014 with a duration of less than 30 seconds), with relatively sufficient semantic objects as reference and multi-view observations for optimization, our algorithm achieves satisfactory results.
However, it can be noted that the advantage of our method is not as obvious on outdoor datasets as on indoor datasets. 
The reasons are two-fold:
One is that the camera moves along a straight path most of the time in outdoor scenes, in which case the quality of dual quadric mapping is not as good as in indoor scenes with the orbit trajectory.
Second, only the ``car'' object can be used as reference for scale estimation in KITTI sequences, and there is no cross-validation with objects of other classes, which makes the overall result prone to bias.

\begin{table}[tbp]
\begin{center}
    \caption{Comparison of Scale Estimation on TUM Dataset}
    \label{tb_rse_tum}
    \begin{tabular}{@{}ccccc@{}}
    \toprule
    \multirow{2}{*}{Seq} & \multicolumn{2}{c}{Sucar's \cite{sucar2017Probabilistic}} & \multicolumn{2}{c}{Ours} \\ \cmidrule(l){2-5} 
                         & RSE(\%)          & std         & RSE(\%)             & std    \\ \midrule
    fr1-desk             & 13.89        & 5.78        & \textbf{4.35}   & 3.51   \\
    fr2-desk             & 5.50         & 2.58        & \textbf{3.63}   & 1.22   \\
    fr3-office           & 9.49         & 3.02        & \textbf{6.27}   & 2.60   \\ \bottomrule
    \end{tabular}
\end{center}
\vspace{-0.3cm}
\end{table}

\begin{table}
    \begin{center}
    \caption{Comparison of Scale Estimation on KITTI Dataset}
    \label{tb_rse_kitti}
    \begin{tabular}{ccccc} 
    \toprule
    \multirow{2}{*}{Seq} & \multicolumn{2}{c}{Sucar's \cite{sucar2017Probabilistic}} & \multicolumn{2}{c}{Ours}  \\ \cmidrule(l){2-5}
                         & RSE(\%)          & std         & RSE(\%)             & std    \\ \midrule
    0000                 & \textbf{11.09} & 2.02     & 17.58         & 2.76      \\
    0001                 & 11.43          & 1.94     & \textbf{9.68} & 2.12      \\
    0007                 & 8.50           & 1.76     & \textbf{7.77} & 2.19      \\
    0009                 & 6.84           & 1.55     & \textbf{4.33} & 2.49      \\
    0011                 & 9.64           & 1.88     & \textbf{9.05} & 2.24      \\
    0014                 & \textbf{9.14}  & 2.02     & 13.44         & 2.12      \\ \bottomrule
    \end{tabular}
\end{center}
\vspace{-0.3cm}
\end{table}


\subsection{Odometery Evaluation Results}
In order to further demonstrate the effect of scale estimation, we evaluated the scale-recovered camera pose error using absolute trajectory error (ATE) proposed in \cite{sturm2012Benchmark}.
As a comparison, we also tested ORB-SLAM2 in monocular, stereo, and RGB-D modes.
Note that no additional scale alignment was performed when calculating the ATE.
The results are shown in \Reftab{tb_traj_tum} and \Reftab{tb_traj_kitti}.
It can be seen that our method significantly reduces the error compared to the monocular mode of ORB-SLAM2.
A centimeter-level accuracy is achieved in the TUM indoor scenes, and the average ATE is about 7 m in the KITTI outdoor scenes, which makes sense in some practical applications such as Augmented Reality and Virtual Reality.
It can also be seen that the object size information is a weaker metric reference compared to the stereo baseline or RGB-D depth measurements due to its relatively large deviation.

\begin{table}[tbp]
    \begin{center}
        \caption{Comparison of Pose Error on TUM Dataset}
        \label{tb_traj_tum}
        \begin{tabular}{ccccccc}
        \toprule
        \multirow{2}{*}{Seq}  & \multicolumn{2}{c}{Ours} & \multicolumn{2}{c}{\makecell[c]{ORB-SLAM2\\(mono)}} & \multicolumn{2}{c}{\makecell[c]{ORB-SLAM2\\(RGB-D)}} \\ \cmidrule(l){2-7} 
                             & ATE(m)          & std         & ATE(m)             & std         & ATE(m)             & std    \\ \midrule
        fr1-desk             & \underline{0.045}           & 0.031       & 0.062   & 0.061   & \textbf{0.016}   & 0.004\\
        fr2-desk             & \underline{0.065}           & 0.021       & 0.914   & 0.085   & \textbf{0.022}   & 0.006\\
        fr3-office           & \underline{0.139}           & 0.054       & 1.210   & 0.019   & \textbf{0.010}   & 0.003\\ \bottomrule
        \end{tabular}
        \begin{tablenotes}
            The \textbf{best} and \underline{second-best} results are highlighted.
        \end{tablenotes}
    \end{center}
    \vspace{-0.3cm}
    \end{table}
    
    \begin{table}
        \begin{center}
        \caption{Comparison of Pose Error on KITTI Dataset}
        \label{tb_traj_kitti}
        \begin{tabular}{ccccccc} 
        \toprule
        \multirow{2}{*}{Seq} & \multicolumn{2}{c}{Ours} & \multicolumn{2}{c}{\makecell[c]{ORB-SLAM2\\(mono)}} & \multicolumn{2}{c}{\makecell[c]{ORB-SLAM2\\(stereo)}} \\ \cmidrule(l){2-7} 
                             & ATE(m)          & std         & ATE(m)             & std         & ATE(m)             & std\\ \midrule
        0000                 & \underline{2.94}            & 0.71        & 16.39               & 0.35        & \textbf{1.44}               & 0.01      \\
        0001                 & \underline{8.17}            & 3.03        & 79.51               & 0.75        & \textbf{3.03}               & 0.04      \\
        0007                 & \underline{7.54}            & 1.22        & 73.29               & 0.27        & \textbf{3.39}               & 0.04      \\
        0009                 & \underline{10.00}           & 3.29        & 163.96              & 0.40        & \textbf{5.97}               & 0.16      \\
        0011                 & \underline{10.79}           & 1.85        & 64.23               & 0.22        & \textbf{1.63}               & 0.05      \\
        0014                 & \underline{1.67}            & 0.27        & 11.89               & 0.28        & \textbf{0.39}               & 0.01     \\ \bottomrule
        \end{tabular}
        \begin{tablenotes}
            The \textbf{best} and \underline{second-best} results are highlighted.
        \end{tablenotes}
    \end{center}
    \vspace{-0.3cm}
    \end{table}

\subsection{Qualitative Results}
\begin{figure*}
    \centering
    \subfigure[]{
        \includegraphics[width=0.23\linewidth]{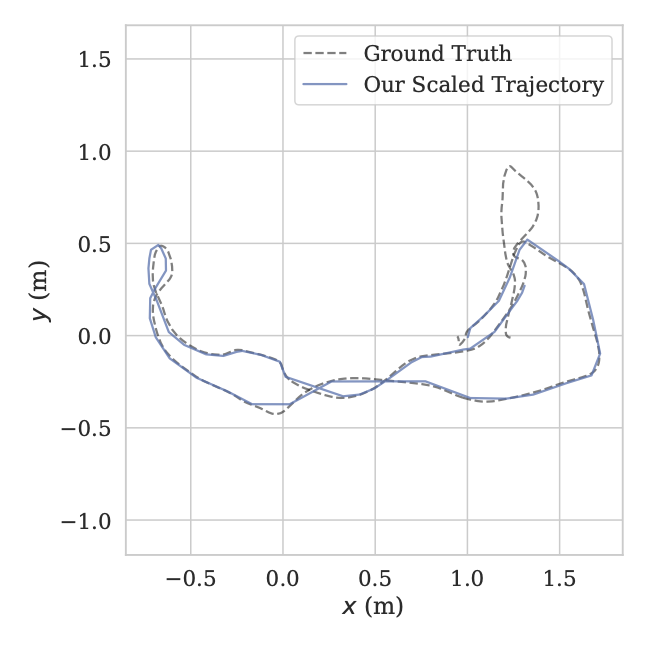} }
    \subfigure[]{
        \includegraphics[width=0.23\linewidth]{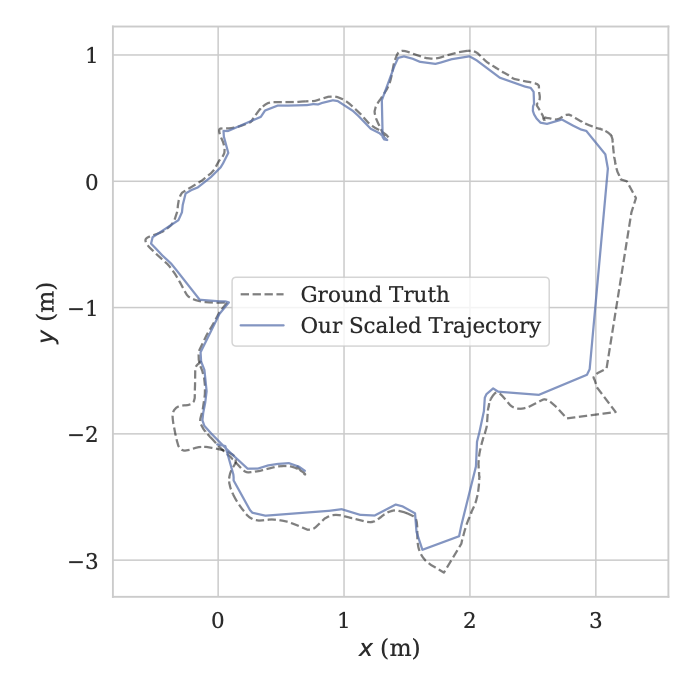} }
    \subfigure[]{
        \includegraphics[width=0.23\linewidth]{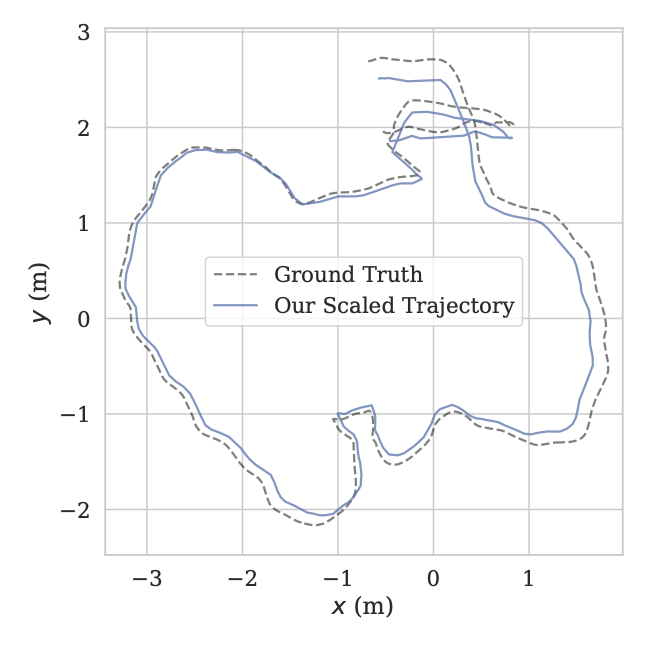} }
    \subfigure[]{
        \includegraphics[width=0.23\linewidth]{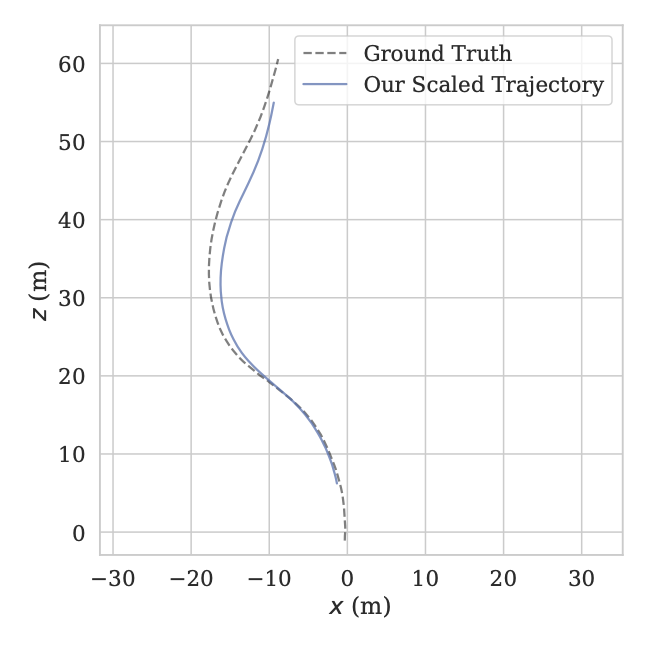} }
    \\
    \subfigure[]{
        \includegraphics[width=0.18\linewidth]{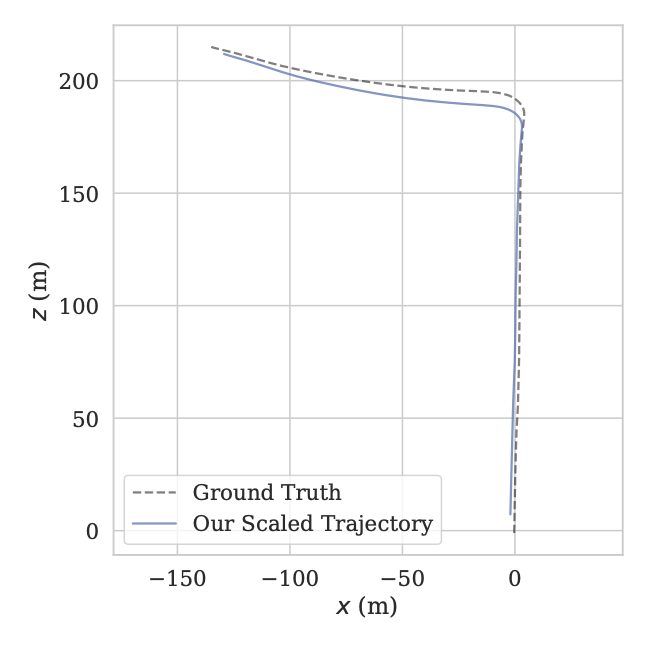} }
    \subfigure[]{
        \includegraphics[width=0.18\linewidth]{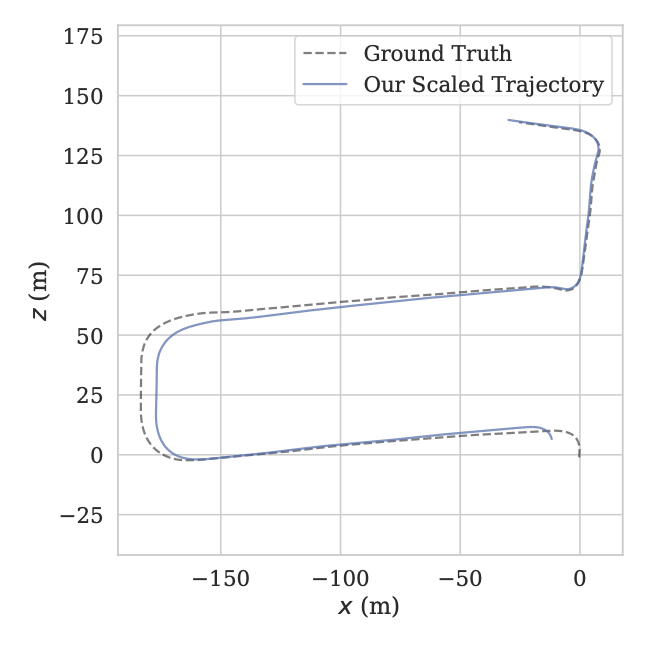} }
    \subfigure[]{
        \includegraphics[width=0.18\linewidth]{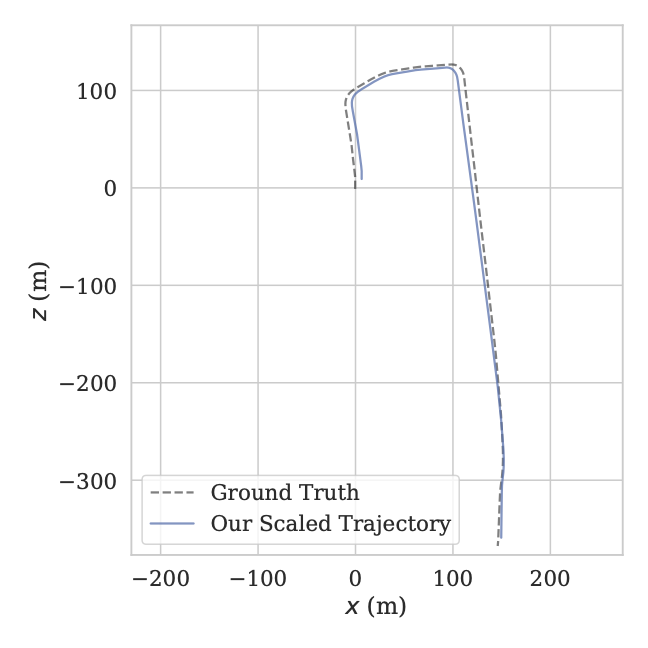} }
    \subfigure[]{
        \includegraphics[width=0.18\linewidth]{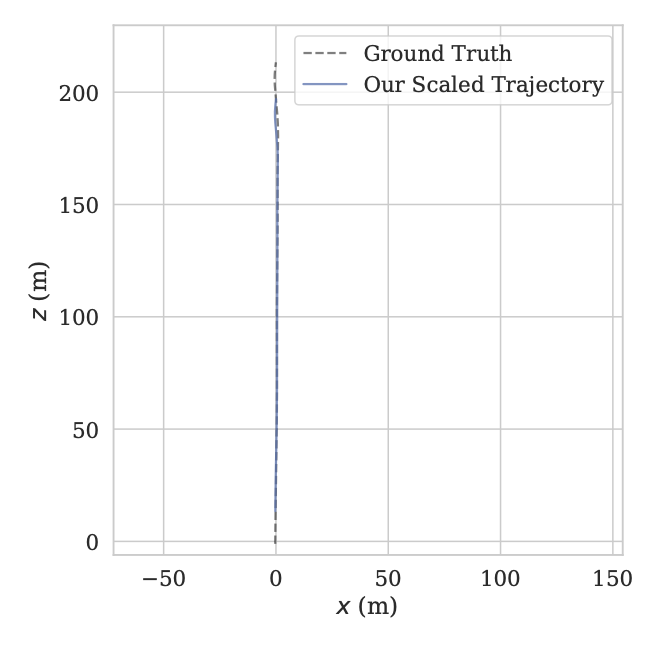} }
    \subfigure[]{
        \includegraphics[width=0.18\linewidth]{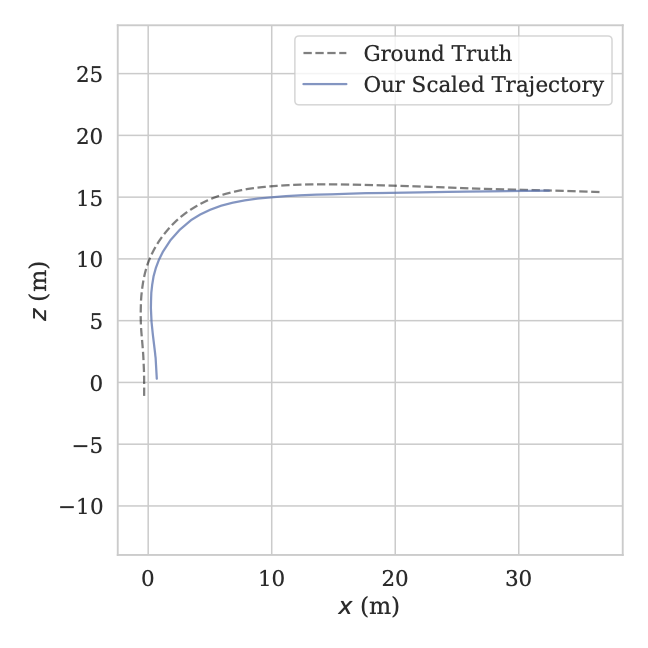} }
    \\
    \caption{Comparisons between the ground truth (gray) and our scaled trajectories (blue). 
    (a), (b) and (c) are results on TUM fr1-desk, fr2-desk and fr3-desk, respectively.
    (d)-(i) are results on KITTI tracking sequences.
    (d) 0000, (e) 0001, (f) 0007, (g) 0009, (h) 0011 and (i) 0014.} 
    \label{fig_qualitative_traj}
    \vspace{-0.3cm}
\end{figure*}

\begin{figure*}
    \centering
    \subfigure[]
    {
        \label{fig_qualitative_proj}
        \includegraphics[width=0.19\linewidth]{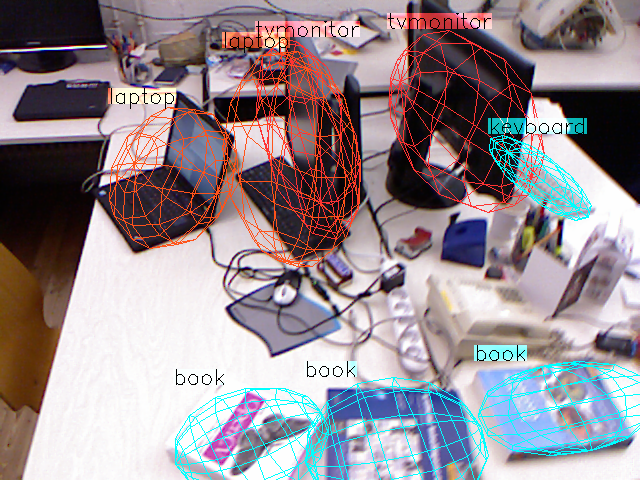}
        \includegraphics[width=0.19\linewidth]{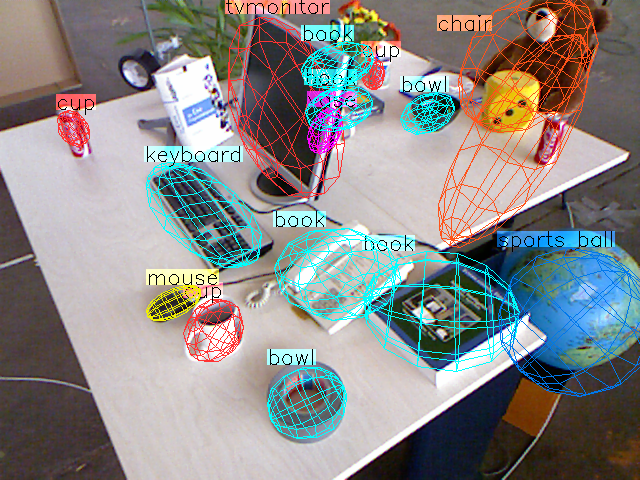}
        \includegraphics[width=0.19\linewidth]{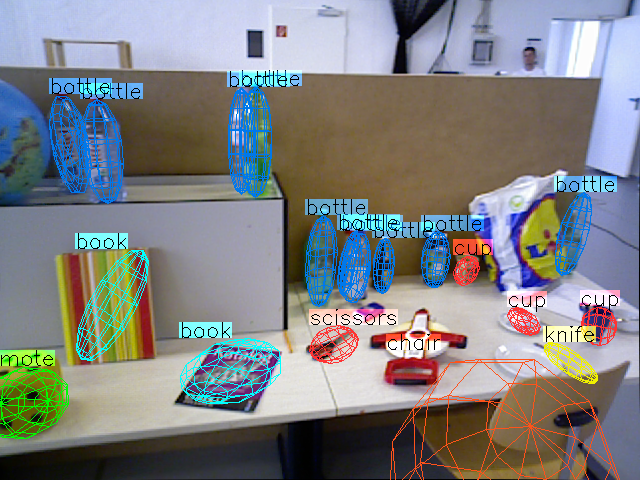}
        \includegraphics[width=0.19\linewidth]{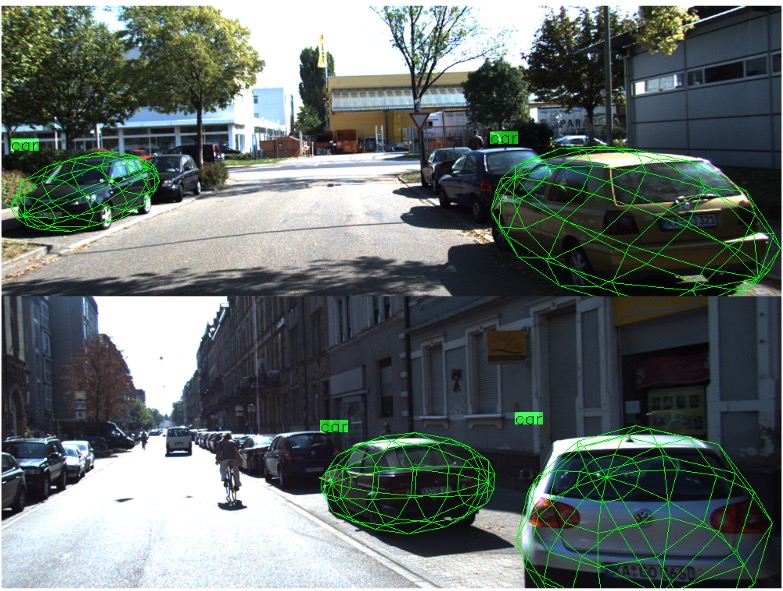}
        \includegraphics[width=0.19\linewidth]{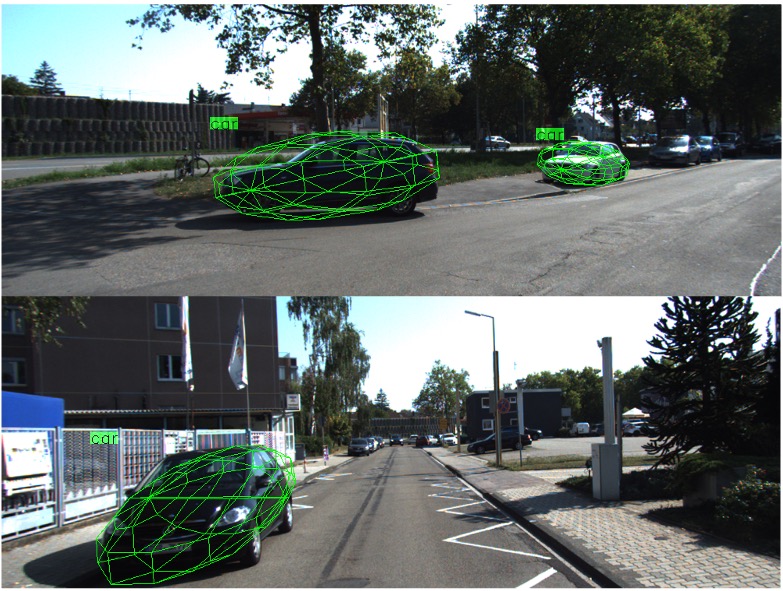}
    }
    \\
    \subfigure[]
    {
        \label{fig_qualitative_quadric}
        \includegraphics[width=0.19\linewidth]{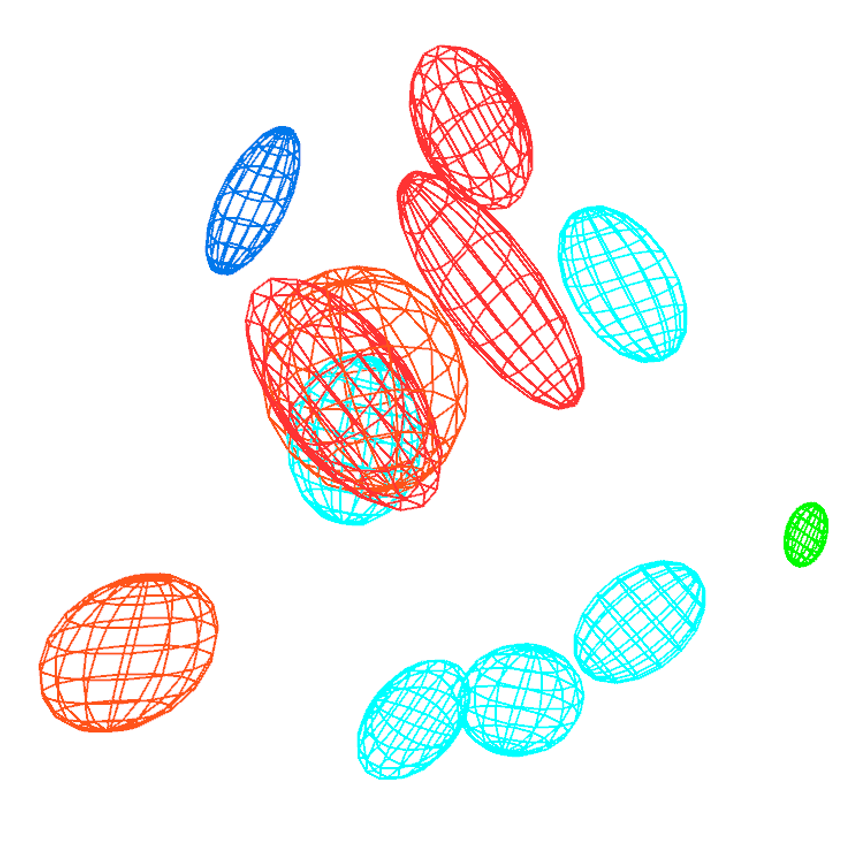}
        \includegraphics[width=0.19\linewidth]{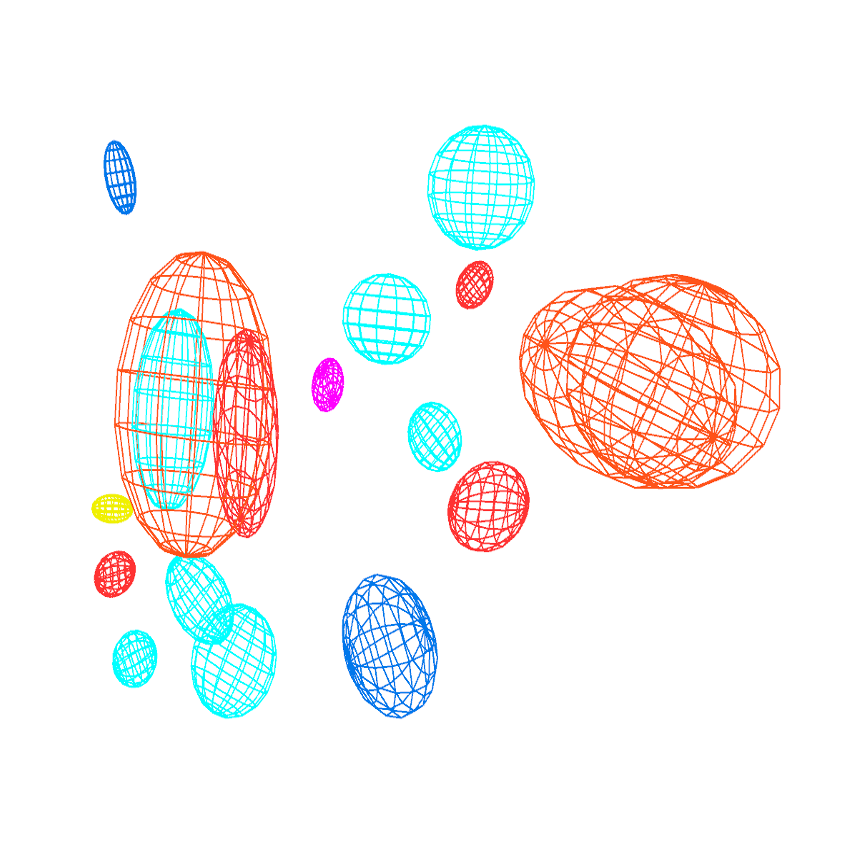}
        \includegraphics[width=0.19\linewidth]{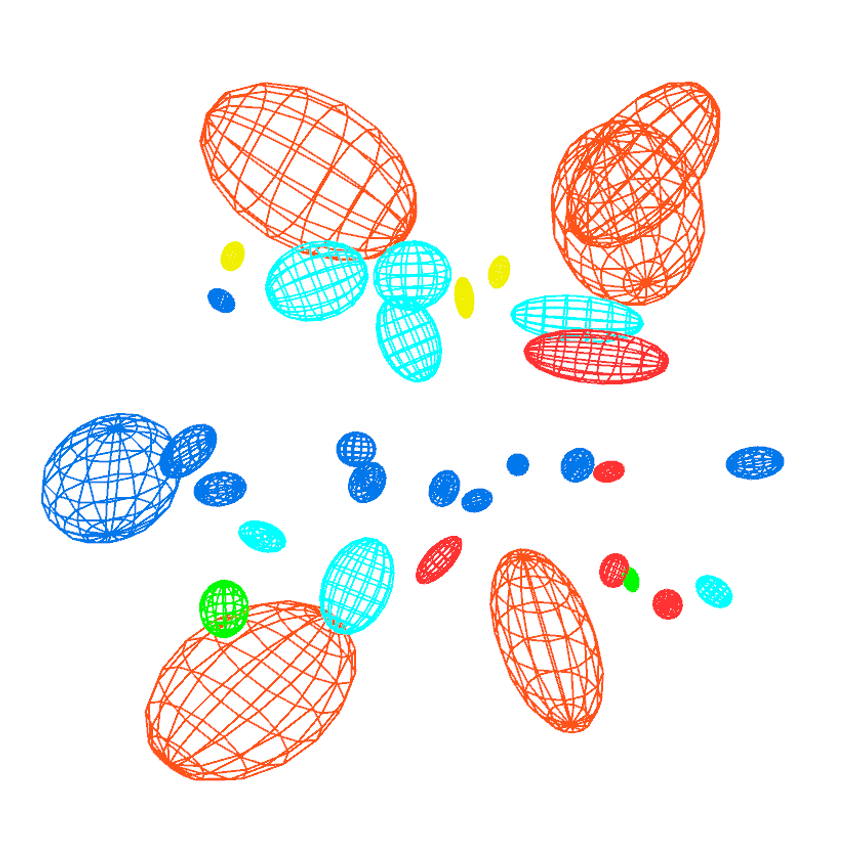}
        \includegraphics[width=0.19\linewidth]{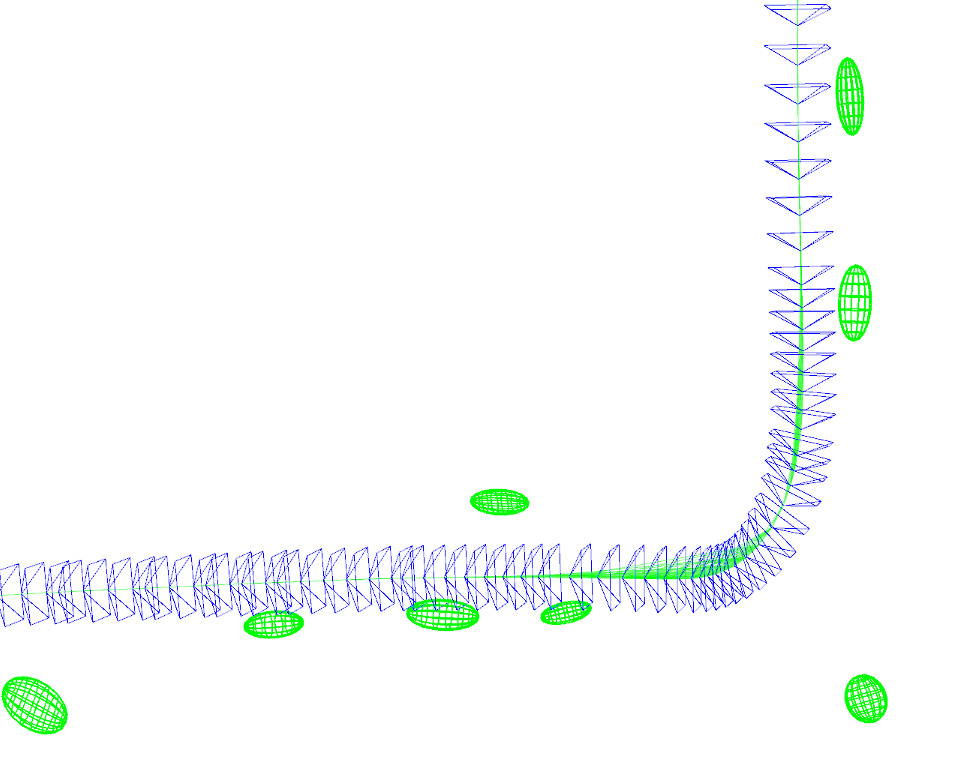}
        \includegraphics[width=0.19\linewidth]{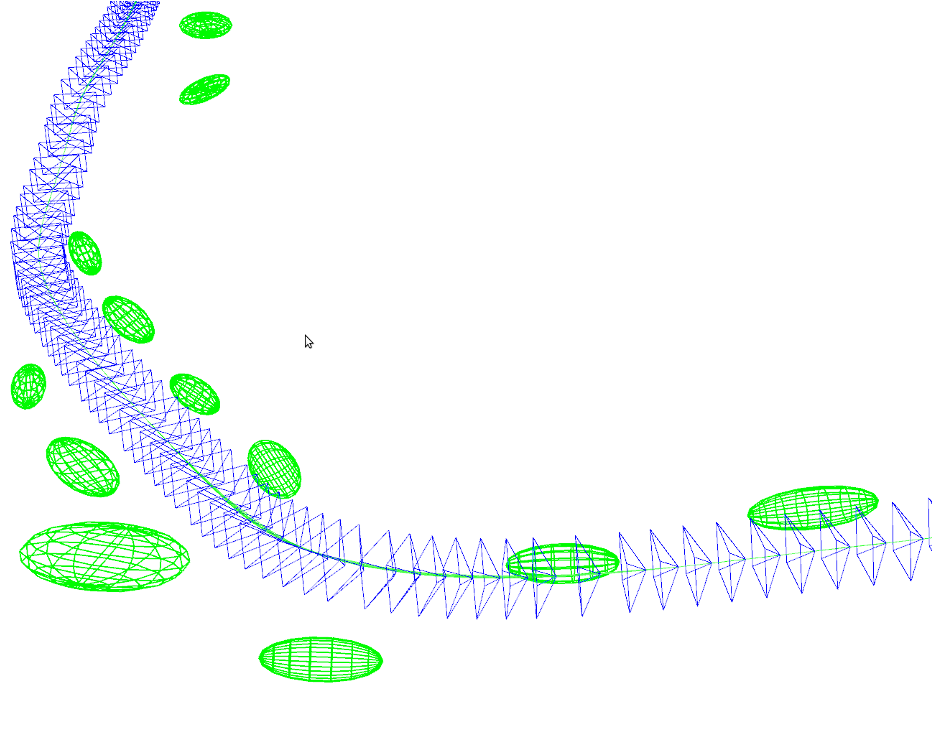}
    }
    \\
    \caption{The qualitative results of scale estimation and semantic mapping. 
    From left to right, columns are results on sequence TUM fr1\_desk, fr2\_desk and fr3\_desk, KITTI tracking 0001 and 0009 respectively.
    (a) The 2-D projection of reconstructed quadrics on the image.
    (b) The object map built by our object SLAM.
    }
    \label{fig_qualitative}
    \vspace{-0.3cm}
\end{figure*}

The qualitative evaluation results of scale estimation and semantic mapping are shown in \Reffig{fig_qualitative_traj} and \Reffig{fig_qualitative} respectively.
\Reffig{fig_qualitative_traj} shows the comparison between the ground truth and the scale-recovered trajectory estimated by our system without further scale alignment.
EVO \cite{grupp2017evo} was used to visualize the trajectories.
\Reffig{fig_qualitative_proj} shows the projection of dual quadrics onto the images.
It can be seen that the reconstructed dual quadrics can capture objects in the scene accurately.
\Reffig{fig_qualitative_quadric} shows that the object-oriented maps built by our system can express the environment well and are understandable with semantic information.


\subsection{Ablation Study}
As explained in \Refsec{sec_scale}, three schemes are designed to enhance our method: outliers elimination, dimensions selection and uncertainty model.
We conduct an ablation study to evaluate their effects on scale estimation.
The results are shown in \Reftab{tb_ablation}.
It can be seen that the relative scale errors are significantly diminished after the process of outliers rejection based on scale consistency.
This is because we get rid of the objects with wrong semantic labels from the false 2-D object detections.
Then, by introducing more reliable dimensions in the scale optimization, the accuracy and robustness of our method are further improved, especially in the KITTI sequences.
Lastly, we adopt the uncertainty model to assign appropriate weights to each error term, and the best performance is achieved.

\begin{table*}[tbp]
    \setlength\tabcolsep{2pt}
    \begin{center}
        \begin{threeparttable}
        \caption{Ablation Study of Different Optional Selecting Results}
        \label{tb_ablation}
        \begin{tabular}{@{}ccccccccccccccccccccc@{}}
        \toprule
        \multirow{2}{*}{out\tnote{1}} & \multirow{2}{*}{dim\tnote{2}} & \multirow{2}{*}{uncer\tnote{3}} & \multicolumn{2}{c}{fr1-desk}  & \multicolumn{2}{c}{fr2-desk}  & \multicolumn{2}{c}{fr3-office} & \multicolumn{2}{c}{0000}  & \multicolumn{2}{c}{0001}  & \multicolumn{2}{c}{0007}  & \multicolumn{2}{c}{0009}  & \multicolumn{2}{c}{0011}  & \multicolumn{2}{c}{0014} \\ \cmidrule(l){4-21} 
                             &                      &                        & RSE(\%)           & std           & RSE(\%)           & std           & RSE(\%)            & std        & RSE(\%)           & std           & RSE(\%)           & std           & RSE(\%)            & std& RSE(\%)           & std           & RSE(\%)           & std           & RSE(\%)            & std   \\ \midrule
                             &                      &                        & 10.81         & 6.56          & 9.10          &3.40          & 14.51          & 4.59           & 29.42 & 4.78 & 15.25 & 2.83 & 22.56 & \textbf{1.89} & 10.87 & 4.56 & 19.89 & 3.14 & 22.36 & 4.53 \\
        $\surd$              &                      &                        & 4.84          & 3.37          & 5.98          & 1.94          & 7.88           & 3.70          & -- & -- & 13.48 & 3.85 & 19.42 & 3.90 & 4.96 & 3.56 & 18.78 & \textbf{2.00} & -- & -- \\
        $\surd$              & $\surd$              &                        & 4.48          & \textbf{3.37} & 3.79          & 1.28          & 6.44           & 2.91          & 17.65 & 2.83 & 10.54 & 2.71 & 8.38 & 2.87 & 4.39 & 2.58 & 9.67 & 2.97 & \textbf{13.42} & \textbf{2.11}\\
        $\surd$              & $\surd$              & $\surd$                & \textbf{4.35} & 3.51          & \textbf{3.63} & \textbf{1.22} & \textbf{6.27}  & \textbf{2.60}  & \textbf{17.28} &\textbf{2.76} &\textbf{9.68}  & \textbf{2.12} & \textbf{7.77} & 2.19 & \textbf{4.33} & \textbf{2.49} &\textbf{9.05}  & 2.24 & 13.44 & 2.12 \\ \bottomrule

    \end{tabular}
    \begin{tablenotes}
        \footnotesize 
        \item[1] Outliers elimination.
        $^2$ Dimensions selection.
        $^3$ Uncertainty model.
        ``--'' means that outliers elimination does not work in sequence 0000 and 0014 because too few dimensions are available before dimensions selection.
      \end{tablenotes}
    \end{threeparttable}
    \end{center}
\end{table*}

\begin{figure}
    \centering
    \includegraphics[width=\linewidth]{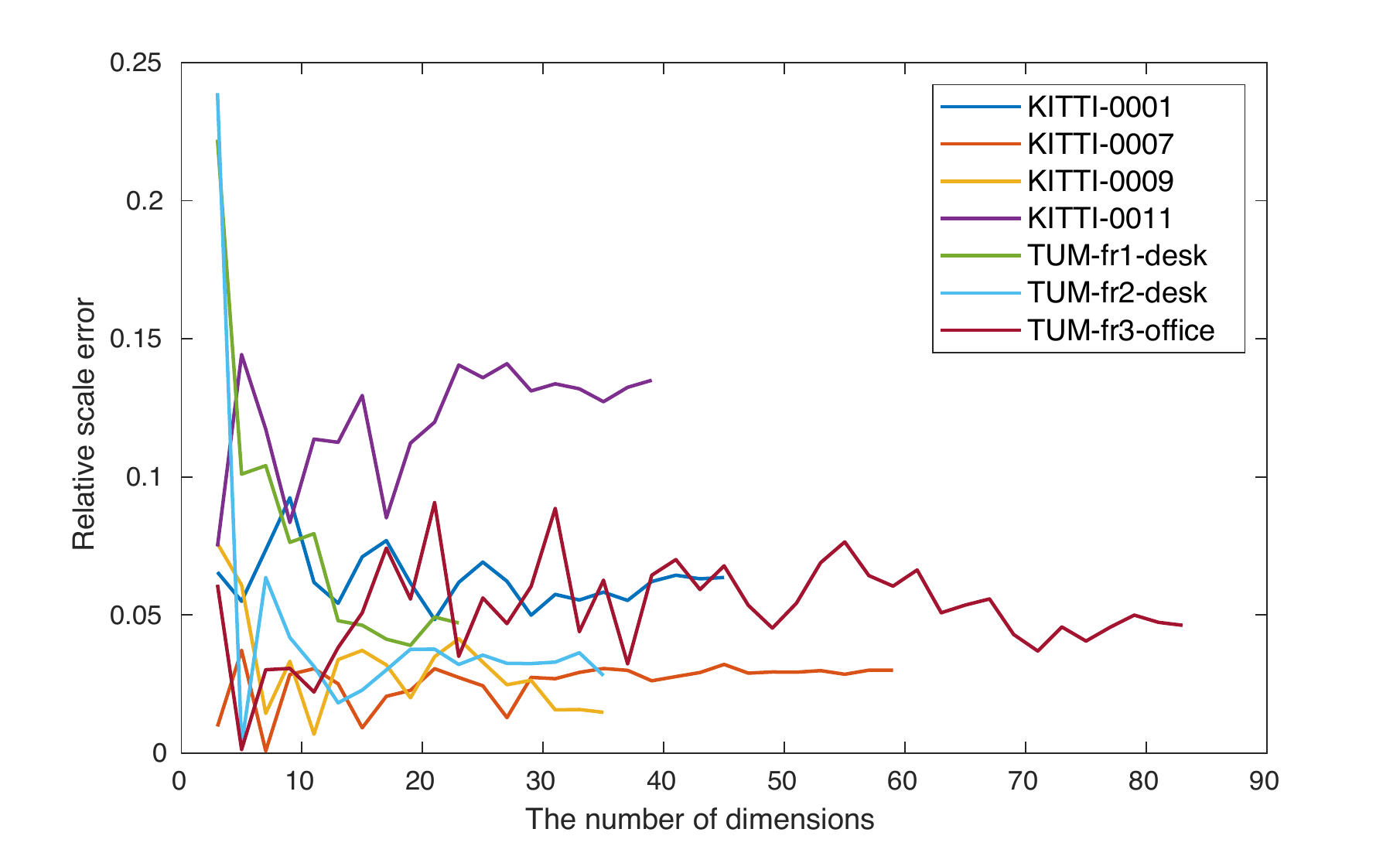}
    \caption{The relation between scale accuracy and the number of available dimensions of objects used in scale estimation.}
    \label{fig_relation}
\end{figure}

Furthermore, we believe that the number of dimensions of objects involved in scale estimation is closely related to the reliability of results.
We run the proposed SLAM system on 7 sequences, recording the number of dimensions and the corresponding RSE.
As shown in \Reffig{fig_relation}, with the increasing number of dimensions, the results of scale estimation tend to be more stable and accurate.


\subsection{Runtime Analysis}
The experiments are carried out on Intel i7-9700K CPU with 3.6GHz and an Nvidia GeForce RTX 2080Ti with 11GB of memory.
On all sequences, the maximum time per frame is around 22ms in the camera tracking thread.
When processing keyframes, object detection and semantic mapping together take about 80ms per image.
Scale estimation module takes less than 1ms.
In summary, our system can run in real time.

\section{CONCLUSION}
In this paper, we present an accurate and robust scale estimation approach that takes the object size priors as the absolute reference.
We develop a monocular object SLAM system to reconstruct objects as dual quadrics to provide reliable dimensions for scale estimation with no resort to assumptions on the gravity direction.
In return, the estimated scale factor can be used to recover the absolute scale of the whole map built by our SLAM system.
Quantitative and qualitative experiments demonstrate the outstanding performance of our method.
In the future, we intend to improve our algorithm to cope with the scale drift problem which often occurs in the large-scale outdoor environment.
Moreover, methods of collecting and modeling prior object size information are also worth exploring.






\bibliographystyle{IEEEtran}
\bibliography{reference}

\begin{thebibliography}{10}
\providecommand{\url}[1]{#1}
\csname url@rmstyle\endcsname
\providecommand{\newblock}{\relax}
\providecommand{\bibinfo}[2]{#2}
\providecommand\BIBentrySTDinterwordspacing{\spaceskip=0pt\relax}
\providecommand\BIBentryALTinterwordstretchfactor{4}
\providecommand\BIBentryALTinterwordspacing{\spaceskip=\fontdimen2\font plus
\BIBentryALTinterwordstretchfactor\fontdimen3\font minus
  \fontdimen4\font\relax}
\providecommand\BIBforeignlanguage[2]{{%
\expandafter\ifx\csname l@#1\endcsname\relax
\typeout{** WARNING: IEEEtran.bst: No hyphenation pattern has been}%
\typeout{** loaded for the language `#1'. Using the pattern for}%
\typeout{** the default language instead.}%
\else
\language=\csname l@#1\endcsname
\fi
#2}}

\bibitem{nicholson2019QuadricSLAM}
L.~Nicholson, M.~Milford, and N.~S{\"u}nderhauf, ``{{QuadricSLAM}}: {{Dual
  Quadrics From Object Detections}} as {{Landmarks}} in {{Object}}-{{Oriented
  SLAM}},'' \emph{IEEE Robotics and Automation Letters}, vol.~4, no.~1, pp.
  1--8, 2019.

\bibitem{yang2019CubeSLAM}
S.~Yang and S.~Scherer, ``{{CubeSLAM}}: {{Monocular}} 3-{{D Object SLAM}},''
  \emph{IEEE Transactions on Robotics}, vol.~35, no.~4, pp. 925--938, 2019.

\bibitem{wu2020EAOSLAM}
Y.~Wu, Y.~Zhang, D.~Zhu, Y.~Feng, S.~Coleman, and D.~Kerr, ``{{EAO}}-{{SLAM}}:
  {{Monocular Semi}}-{{Dense Object SLAM Based}} on {{Ensemble Data
  Association}},'' in \emph{2020 {{IEEE}}/{{RSJ International Conference}} on
  {{Intelligent Robots}} and {{Systems}} ({{IROS}})}.\hskip 1em plus 0.5em
  minus 0.4em\relax {IEEE}, 2020, pp. 4966--4973.

\bibitem{konkle2012RealWorld}
T.~Konkle and A.~Oliva, ``A {{Real}}-{{World Size Organization}} of {{Object
  Responses}} in {{Occipitotemporal Cortex}},'' \emph{Neuron}, vol.~74, no.~6,
  pp. 1114--1124, 2012.

\bibitem{sucar2017Probabilistic}
E.~Sucar and J.-B. Hayet, ``Probabilistic {{Global Scale Estimation}} for
  {{MonoSLAM Based}} on {{Generic Object Detection}},'' in \emph{2017 {{IEEE
  Conference}} on {{Computer Vision}} and {{Pattern Recognition Workshops}}
  ({{CVPRW}})}.\hskip 1em plus 0.5em minus 0.4em\relax {IEEE}, 2017, pp.
  988--996.

\bibitem{sucar2018Bayesian}
E.~Sucar and J.-B. Hayet, ``Bayesian {{Scale Estimation}} for {{Monocular SLAM
  Based}} on {{Generic Object Detection}} for {{Correcting Scale Drift}},'' in
  \emph{2018 {{IEEE International Conference}} on {{Robotics}} and
  {{Automation}} ({{ICRA}})}.\hskip 1em plus 0.5em minus 0.4em\relax {IEEE},
  2018, pp. 5152--5158.

\bibitem{Hartley2003}
R.~Hartley and A.~Zisserman, \emph{Multiple View Geometry in Computer
  Vision}.\hskip 1em plus 0.5em minus 0.4em\relax Cambridge, U.K.: Cambridge
  Univ. Press, 2004.

\bibitem{chen2021Robust}
S.~Chen, S.~Song, J.~Zhao, T.~Feng, C.~Ye, L.~Xiong, and D.~Li, ``Robust {{Dual
  Quadric Initialization}} for {{Forward}}-{{Translating Camera Movements}},''
  \emph{IEEE Robotics and Automation Letters}, vol.~6, no.~3, pp. 4712--4719,
  2021.

\bibitem{zhang2020Scaleaware}
S.~Zhang, X.~Li, Y.~Liu, and H.~Fu, ``Scale-aware {{Insertion}} of {{Virtual
  Objects}} in {{Monocular Videos}},'' in \emph{2020 {{IEEE International
  Symposium}} on {{Mixed}} and {{Augmented Reality}} ({{ISMAR}})}.\hskip 1em
  plus 0.5em minus 0.4em\relax {IEEE}, 2020, pp. 36--44.

\bibitem{nutzi2011Fusion}
G.~N{\"u}tzi, S.~Weiss, D.~Scaramuzza, and R.~Siegwart, ``Fusion of {{IMU}} and
  {{Vision}} for {{Absolute Scale Estimation}} in {{Monocular SLAM}},''
  \emph{Journal of Intelligent \& Robotic Systems}, vol.~61, no.~1, pp.
  287--299, 2011.

\bibitem{zhang2018Scale}
Z.~Zhang, R.~Zhao, E.~Liu, K.~Yan, and Y.~Ma, ``Scale {{Estimation}} and
  {{Correction}} of the {{Monocular Simultaneous Localization}} and {{Mapping}}
  ({{SLAM}}) {{Based}} on {{Fusion}} of {{1D Laser Range Finder}} and {{Vision
  Data}},'' \emph{Sensors}, vol.~18, no.~6, p. 1948, 2018.

\bibitem{song2014Robust}
S.~Song and M.~Chandraker, ``Robust {{Scale Estimation}} in {{Real}}-{{Time
  Monocular SFM}} for {{Autonomous Driving}},'' in \emph{2014 {{IEEE
  Conference}} on {{Computer Vision}} and {{Pattern Recognition}}}.\hskip 1em
  plus 0.5em minus 0.4em\relax {IEEE}, 2014, pp. 1566--1573.

\bibitem{zhou2016Reliable}
{Dingfu Zhou}, Y.~Dai, and {Hongdong Li}, ``Reliable scale estimation and
  correction for monocular {{Visual Odometry}},'' in \emph{2016 {{IEEE
  Intelligent Vehicles Symposium}} ({{IV}})}.\hskip 1em plus 0.5em minus
  0.4em\relax {IEEE}, 2016, pp. 490--495.

\bibitem{wang2018Monocular}
X.~Wang, H.~Zhang, X.~Yin, M.~Du, and Q.~Chen, ``Monocular {{Visual Odometry
  Scale Recovery Using Geometrical Constraint}},'' in \emph{2018 {{IEEE
  International Conference}} on {{Robotics}} and {{Automation}}
  ({{ICRA}})}.\hskip 1em plus 0.5em minus 0.4em\relax {IEEE}, 2018, pp.
  988--995.

\bibitem{yin2017Scale}
X.~Yin, X.~Wang, X.~Du, and Q.~Chen, ``Scale {{Recovery}} for {{Monocular
  Visual Odometry Using Depth Estimated}} with {{Deep Convolutional Neural
  Fields}},'' in \emph{2017 {{IEEE International Conference}} on {{Computer
  Vision}} ({{ICCV}})}.\hskip 1em plus 0.5em minus 0.4em\relax {IEEE}, 2017,
  pp. 5871--5879.

\bibitem{yang2018Deep}
N.~Yang, R.~Wang, J.~St{\"u}ckler, and D.~Cremers, ``Deep virtual stereo
  odometry: {{Leveraging}} deep depth prediction for monocular direct sparse
  odometry,'' in \emph{Proceedings of the {{European Conference}} on {{Computer
  Vision}} ({{ECCV}})}.\hskip 1em plus 0.5em minus 0.4em\relax {Springer
  International Publishing}, 2018, pp. 835--852.

\bibitem{wang2017Stereo}
R.~Wang, M.~Schworer, and D.~Cremers, ``Stereo {{DSO}}: {{Large}}-{{Scale
  Direct Sparse Visual Odometry With Stereo Cameras}},'' in \emph{Proceedings
  of the {{IEEE International Conference}} on {{Computer Vision}}}.\hskip 1em
  plus 0.5em minus 0.4em\relax {IEEE}, 2017, pp. 3903--3911.

\bibitem{greene2020Metrically}
W.~N. Greene and N.~Roy, ``Metrically-{{Scaled Monocular SLAM}} using {{Learned
  Scale Factors}},'' in \emph{2020 {{IEEE International Conference}} on
  {{Robotics}} and {{Automation}} ({{ICRA}})}.\hskip 1em plus 0.5em minus
  0.4em\relax {IEEE}, 2020, pp. 43--50.

\bibitem{salas2013SLAM}
R.~F. {Salas-Moreno}, R.~A. Newcombe, H.~Strasdat, P.~H. Kelly, and A.~J.
  Davison, ``{{SLAM}}++: {{Simultaneous Localisation}} and {{Mapping}} at the
  {{Level}} of {{Objects}},'' in \emph{2013 {{IEEE Conference}} on {{Computer
  Vision}} and {{Pattern Recognition}}}.\hskip 1em plus 0.5em minus 0.4em\relax
  {IEEE}, 2013, pp. 1352--1359.

\bibitem{galvez2016Real}
D.~{G{\'a}lvez-L{\'o}pez}, M.~Salas, J.~D. Tard{\'o}s, and J.~M.~M. Montiel,
  ``Real-time monocular object {{SLAM}},'' \emph{Robotics and Autonomous
  Systems}, vol.~75, pp. 435--449, 2016.

\bibitem{frost2018Recovering}
D.~Frost, V.~Prisacariu, and D.~Murray, ``Recovering {{Stable Scale}} in
  {{Monocular SLAM Using Object}}-{{Supplemented Bundle Adjustment}},''
  \emph{IEEE Transactions on Robotics}, vol.~34, no.~3, pp. 736--747, 2018.

\bibitem{weinmann2014Semantic}
M.~Weinmann, B.~Jutzi, and C.~Mallet, ``Semantic {{3D}} scene interpretation:
  {{A}} framework combining optimal neighborhood size selection with relevant
  features,'' \emph{ISPRS Annals of the Photogrammetry, Remote Sensing and
  Spatial Information Sciences}, vol. II-3, pp. 181--188, 2014.

\bibitem{kummerle2011G2o}
R.~K{\"u}mmerle, G.~Grisetti, H.~Strasdat, K.~Konolige, and W.~Burgard, ``G2o:
  {{A}} general framework for graph optimization,'' in \emph{2011 {{IEEE
  International Conference}} on {{Robotics}} and {{Automation}}}.\hskip 1em
  plus 0.5em minus 0.4em\relax {IEEE}, 2011, pp. 3607--3613.

\bibitem{mur-artal2017ORBSLAM2}
R.~{Mur-Artal} and J.~D. Tard{\'o}s, ``{{ORB}}-{{SLAM2}}: {{An Open}}-{{Source
  SLAM System}} for {{Monocular}}, {{Stereo}}, and {{RGB}}-{{D Cameras}},''
  \emph{IEEE Transactions on Robotics}, vol.~33, no.~5, pp. 1255--1262, 2017.

\bibitem{gottschalk2000collision}
S.~A. Gottschalk, ``Collision queries using oriented bounding boxes,''
  Ph.{{D}}., The University of North Carolina at Chapel Hill, 2000.

\bibitem{ok2019Robust}
K.~Ok, K.~Liu, K.~Frey, J.~P. How, and N.~Roy, ``Robust {{Object}}-based
  {{SLAM}} for {{High}}-speed {{Autonomous Navigation}},'' in \emph{2019
  {{International Conference}} on {{Robotics}} and {{Automation}}
  ({{ICRA}})}.\hskip 1em plus 0.5em minus 0.4em\relax {IEEE}, 2019, pp.
  669--675.

\bibitem{sturm2012Benchmark}
J.~Sturm, N.~Engelhard, F.~Endres, W.~Burgard, and D.~Cremers, ``A benchmark
  for the evaluation of {{RGB-D SLAM}} systems,'' in \emph{2012 {{IEEE}}/{{RSJ
  International Conference}} on {{Intelligent Robots}} and {{Systems}}}.\hskip
  1em plus 0.5em minus 0.4em\relax {IEEE}, 2012, pp. 573--580.

\bibitem{geiger2012Are}
A.~Geiger, P.~Lenz, and R.~Urtasun, ``Are we ready for autonomous driving?
  {{The KITTI}} vision benchmark suite,'' in \emph{2012 {{IEEE Conference}} on
  {{Computer Vision}} and {{Pattern Recognition}}}.\hskip 1em plus 0.5em minus
  0.4em\relax {IEEE}, 2012, pp. 3354--3361.

\bibitem{redmon2018YOLOv3}
J.~{Redmon} and A.~{Farhadi}, ``{{YOLOv3}}: {{An Incremental Improvement}},''
  \emph{arXiv e-prints}, 2018.

\bibitem{grupp2017evo}
M.~Grupp, ``evo: Python package for the evaluation of odometry and slam.''
  \url{https://github.com/MichaelGrupp/evo}, 2017.

\end{thebibliography}
\end{document}